\documentclass[10pt,journal,compsoc]{IEEEtran}
\usepackage{amsmath,amssymb,amsfonts}
\usepackage{graphicx}
\usepackage{ragged2e}
\usepackage{textcomp}

\usepackage[ruled,vlined]{algorithm2e}
\usepackage{caption, subcaption}
\usepackage{todonotes}
\usepackage[para]{footmisc}
\usepackage{booktabs}
\usepackage{xcolor,colortbl}
\usepackage{hyperref}
\usepackage{multirow}
\usepackage{cleveref}
\usepackage{rotating}
\usepackage{wrapfig}
\usepackage[noadjust]{cite}
\usepackage{paralist}
\usepackage{listings}
\usepackage{lipsum}
\usepackage{amsthm}
\usepackage[para]{footmisc}

\begin{document}

\title{Catch Me If You Can: Semi-supervised Graph Learning for Spotting Money Laundering}

\author{Md. Rezaul Karim, Felix Hermsen, Sisay Adugna Chala, Paola de Perthuis, and Avikarsha Mandal

\thanks{Md. Rezaul Karim is with Fraunhofer FIT \& RWTH Aachen University\\ 
        Felix Hermsen, Sisay A. Chala, \& Avikarsha Mandal are with Fraunhofer FIT \\
        Paola de Perthuis is with Cosmian, France. }% <-this % stops a space
}

\markboth{}%
{Karim and Mandal \MakeLowercase{\textit{et al.}}: Interpreting Black-box Machine Learning Models for high-dimensional Datasets}

\maketitle

\begin{abstract}
\justifying
     % Money laundering is the process where criminals use financial services to move massive amounts of illegal money to untraceable destination and integrate them into legitimate financial systems. Therefore, it is of uttermost importance for any country to identify such activities accurately and reliably in order to enforce AML. From given graphs of the money transfers between accounts of a bank, existing approaches attempted to detect money laundering. Despite of tremendous  efforts to anti-money laundering~(AML) only a tiny fraction of illicit activities is prevented. In particular, some approaches employ structural and behavioral dynamics of dense subgraph detection thereby not taking into consideration that money laundering involves high-volume flows of funds through chains of bank accounts. Some approaches model the transactions in the form of multipartite graphs to detect the complete flow of money from source to destination. These approaches, yielding lower detection accuracy found to be less reliability. In this paper, we employed semi-supervised graph learning techniques on graphs of financial transactions in order to identify nodes involved in potential money laundering. Experimental results show that our approach outperforms state-of-the-art approaches in accurately detecting when tested on large graphs. 
     Money laundering is the process where criminals use financial services to move massive amounts of illegal money to untraceable destinations and integrate them into legitimate financial systems. It is very crucial to identify such activities accurately and reliably in order to enforce anti-money laundering~(AML). Despite tremendous efforts to AML only a tiny fraction of illicit activities is prevented. From a given graph of money transfers between accounts of a bank, existing approaches attempted to detect money laundering. 
     In particular, some approaches employ structural and behavioural dynamics of dense subgraph detection thereby not taking into consideration that money laundering involves high-volume flows of funds through chains of bank accounts. Some approaches model the transactions in the form of multipartite graphs to detect the complete flow of money from source to destination. 
     However, existing approaches yield lower detection accuracy, making them less reliable. In this paper, we employ semi-supervised graph learning techniques on graphs of financial transactions in order to identify nodes involved in potential money laundering. Experimental results\footnote{\scriptsize{Codes to appear at: \url{https://github.com/AwesomeDeepAI/Graph-based-money-laundering-detection}}} suggest that our approach can sport money laundering from real and synthetic transaction graphs. 
\end{abstract}

\begin{IEEEkeywords} Money laundering, Graph embedding, Machine learning. \end{IEEEkeywords}

\section{Introduction}
Money laundering is an alarming problem globally. It causes approximately 1.6 Trillion USD corresponding to 2.7\% of the global GDP is laundered every year~\cite{frumerie2021money, FACTI_Panel}.  Criminals involved in money laundering activities, often hide the original sources of illegal money by using the funds in casinos or real estate purchases, or by overvaluing legitimate invoices. There are many ways of money laundering, yet generally, it is composed of three primary steps: \emph{placement}, \emph{layering}, and \emph{integration}. %, as shown in \cref{fig:money_laundering_life_cycle}. 
Placement is about introducing dirty money into existing financial systems. Layering is the process of carrying out complex transactions to hide the source of the funds. The integration is about withdrawing the proceeds from a destination bank account before using the fund for legitimate activities. 

Anti-money laundering~(AML) is the task of preventing criminals from moving illicit funds through the financial system. AML is often perceived through regulatory compliance since the burden of forensic analysis falls primarily on financial institutions that are responsible to comply with Know Your Customer~(KYC) standards, monitoring transactions, shutting down or restricting accounts deemed suspect, and submitting timely Suspicious Activities Reports~(SARs) to law enforcement agencies. These are typically carried out in a five-step process: 

\begin{enumerate}
    \item introducing a compliance organization within the company including formal AML training for employees. 
    \item emphasising and execution of KYC onboarding and profile maintenance procedures. 
    \item account activity oversight and constraints via transaction monitoring systems~(TMS). 
    \item manual review of flagged accounts and transactions. 
    \item filing of SARs to law enforcement and corresponding restrictive action against suspect accounts. 
\end{enumerate}

Transaction monitoring systems are predominantly rules-based thresholding protocols tuned for the volume and velocity of transactions with tiered escalation procedures. Thorough analysis is carried out using sophisticated techniques to determine whether or not a SAR need to be filed and the account in question suspended. 
%As shown in \cref{fig:ml_topologies}, 
There are two main topologies to represent how the money-laundering is carried out: The first topology involves one sender account and one receiver account, and many intermediary accounts forming a column. The funds are divided by the sender among intermediary accounts, which then resend the funds to the receiver account. The second topology consists of one sender and one receiver and a row of intermediary accounts. The intermediary accounts pass the entire amount\footnote{\scriptsize{Sometimes minus the optional commission.}} of funds to the next account in the row until they are received at the receiver account. 

A money transfer or transaction graph can be constructed such that a single account is represented as a vertex and a single transaction between two accounts is represented as an edge. Besides, a group of accounts can be represented as a vertex~(under a holding company or inferred to shareowners), while an edge can represent the aggregate transaction volume with a neighbouring node over a period of time. Further, a node entity might be a single account or a set of associated accounts in AML transaction monitoring.

% \begin{figure}[h]
%     \centering
%     \includegraphics[width=0.5\textwidth]{images/aml.png}
%     \caption{Money laundering cycle}
%     \label{fig:money_laundering_life_cycle}
% \end{figure}

Identifying money launders from such a graph is very challenging. Automatically or semi-automatically spotting culprits\footnote{\scriptsize{That could potentially involve money laundering activities.}} from a massive graph data by mapping billions of edges between millions of entities~(nodes) needs a scalable and efficient method. Nevertheless, criminals can be quite sophisticated in masking the true nature of their transactions with complicated account layering\footnote{\scriptsize{In order to confuse AML tools and methods.}} or multi-hop transactions, making the identification of money laundering extremely difficult and computationally complex problem. Consequently, using existing methods or tools, only a tiny fraction of illicit activities is prevented despite of tremendous efforts to AML, while penalties for failed AML compliance are severe. Therefore, it is of uttermost importance for any country or financial organization to identify such activities accurately and reliably in order to enforce AML. 

Further, recent approaches that employed non-graph-based models mostly benefit from using additional features generated by graph-based models such as graph neural networks~(GNNs)~\cite{wu2020comprehensive}. These inspired us to combine the power of graph analytics with tree-based ensemble models, thereby modelling both spatial and temporal information in a large-scale graph, in semi-supervised learning settings~\cite{Kipf2017}. We hypothesize that similar to network analysis that involves predictions over nodes and edges~(e.g., predicting most probable labels of nodes in a network), nodes showing distinct characteristics from regular nodes can be classified as potential money launders in a transaction graph, by modelling it as a node classification problem. In the end, we employ a semi-supervised graph learning technique on the graph of financial transactions in order to identify nodes involved in potential money laundering. 

%The rest of the paper is structured as follows: \Cref{sec:rw} critically reviews related works. \Cref{sec:methods} describes our proposed approach. \Cref{sec:exp} reports experiment results, including a comparative analysis with baseline models. \Cref{sec:con} summarizes this research with potential limitations and points to some possible outlooks before concluding the paper.

\begin{figure*}
	\centering
	\begin{subfigure}{\linewidth}
		\centering
		\includegraphics[width=0.6\textwidth]{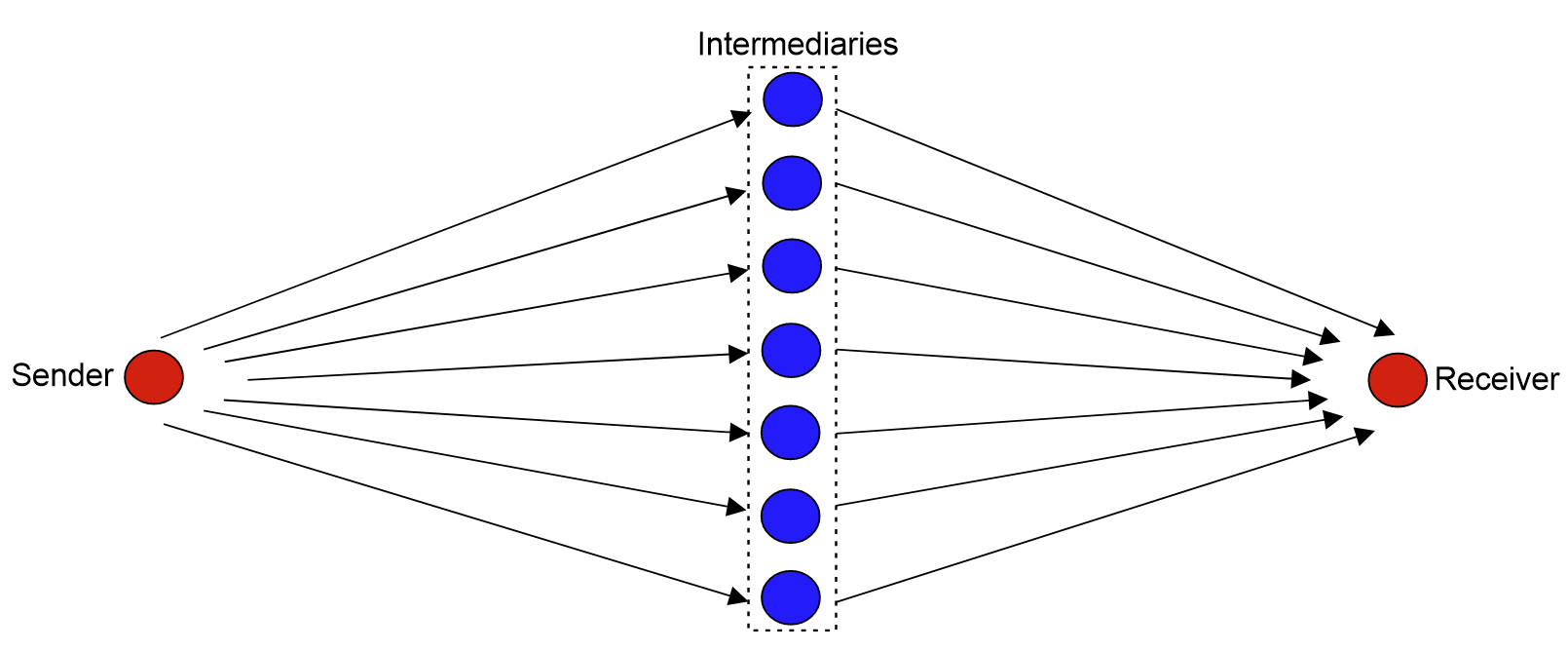}
		%\caption{Probing a black-box model}
        \label{fig:probing}
	\end{subfigure}
	\begin{subfigure}{\linewidth}
		\centering
		\includegraphics[width=0.6\textwidth]{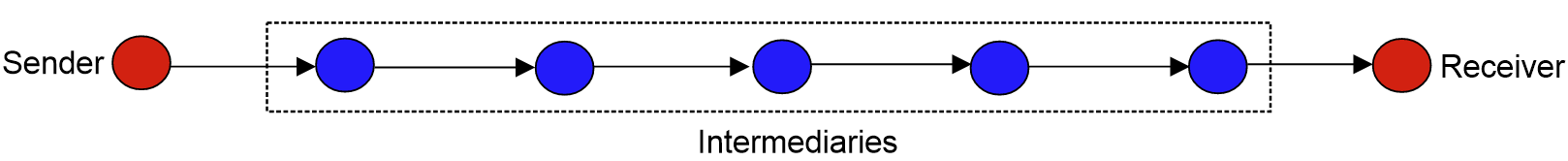}
		%\caption{Interpreting black-box model with perturbing}
        \label{fig:purturbing}
	\end{subfigure}
	\caption{Two common topologies of money laundering networks~(recreated based on literature~\cite{Soltani2016new})} 
	\label{fig:ml_topologies}
\end{figure*}

\section{Related Works}\label{sec:rw}
Several methods have been proposed to accurately identify AML activities~\cite{li2020flowscope}. The earliest approaches predominantly relied on rule-based classification. For example, Rajput et al.~\cite{rajput2014ontology} developed an ontology-based expert system to detect suspicious transactions. However, rule-based algorithms are easy to be evaded by fraudsters~\cite{li2020flowscope}. Therefore, graph and ML-based approaches have emerged. Michalak et al.~\cite{michalak2011graph} used fuzzy matching to capture subgraphs that are more likely to contain suspicious accounts involved in fraudulent activities. A few other approaches tried to assess if the capital flow is involved in money laundering activities using radial basis function~(RBF) neural networks calculating w.r.t time-to-times~\cite{li2020flowscope}. From a graph of money transfers between accounts, existing approaches attempted to detect money laundering activities by employing a variety of methods starting from simple logistic regression~(LR), support vector machines~(SVM)~\cite{tang2005developing}, random forest~(RF), and multilayer perceptron~(MLP) to more sophisticated approach that are based on GNNs~\cite{wu2020comprehensive}. Some earlier approaches~\cite{awasthi2012clustering,le2010application} employed a clustering-based method to detect money laundering activities by grouping transactions into clusters. Soltani et al.~\cite{Soltani2016new} proposed a money laundering detection algorithm. This structural similarity-based approach finds pairs of transactions with common attributes and behaviours that potentially involve money laundering activities. 

Money laundering usually involves high-volume flows of funds through chains of bank accounts~\cite{li2020flowscope}. However, approaches that employ structural and behavioural dynamics of dense subgraph detection do not take it into account~\cite{li2020flowscope}. Methods that do not perform flow tracking, yield lower detection accuracy and hence cannot provide theoretical guarantees, while the flow across multiple nodes is important for accuracy and robustness against camouflage in the money laundering activities~\cite{hooi2016fraudar}. However, real-world data is typically either unlabeled, or has noisy, or sparse labels, whereas these algorithms detect money laundering activities in a supervised manner, suffering from highly skewed labels and limited adaptability. Further, money laundering activities often involve cash flow relationships between entities, i.e., network structures~\cite{li2020flowscope}. Therefore, some approaches model the transactions in the form of multipartite graphs to detect the complete flow of money from source to destination~\cite{li2020flowscope} in an unsupervised manner based on graphs. In particular, FlowScope~\cite{li2020flowscope} is a flow-based approach for detecting money laundering behaviour to detect the chains of transactions. 

A knowledge graph~(KG) can be formed in a similar fashion, where nodes represent entities and edges represent binary relations between those entities~\cite{hogan2020knowledge}. More formally, $\mathcal{G} = \{E,R,T\}$, where $\mathcal{G}$ is a labelled and directed multi-graph, and $E, R, T$ are the sets of entities, relations, and triples, respectively. Each triple\footnote{\scriptsize{The notation $<head,relation,tail>$ is called resource description framework~(RDF) - a W3C standard for serializing data.}} in the KG can be formalized as $(h,r,t) \in T$, where $h \in E$ is the head node, $t \in E$ is the tail node, and $r \in R$ is the edge connecting $h$ and $t$~(i.e., the relation $r$ holds between $h$ and $t$)~\cite{hogan2020knowledge}. Recently, graph analytics techniques on such KGs have emerged as an increasingly important means for AML analysis. In particular, GNN architectures such as graph convolutional networks~(GCN)~\cite{Kipf2017}, GraphSAGE~\cite{GraphSAGE}, and FastGCN~\cite{Chen2018} have emerged~\cite{Soltani2016new}, showing efficiency and scalability in graph-based representation learning~\cite{Chen2018,Ying2018}. The reason is that in low-labelled dataset scenarios, unsupervised techniques can learn low-dimensional meaningful representations of nodes by leveraging the graph structure and features. An unsupervised GE model learns embeddings of unlabeled graph nodes. Further, literature has shown that the inclusion of non-local information -- specifically information about the neighbours of a centre node always improves the performance of each model, non-graph-based models mostly benefit from using additional features given by a GE model. This makes graph learning-based approaches quite promising due to their ability to understand complex patterns and feature discovery capabilities. 

Generating quality feature vectors using appropriate graph embedding~(GE) methods play a significant role in any downstream learning task. Numerous GE methods have been proposed to date~\cite{wu2020comprehensive}. A GE technique aims to map a KG into a dense, low-, feature space, which is capable of preserving as much structure and property information of the graph as possible and aiding in calculations of the entities and relations\cite{dai2020survey}. A GE model consists of three steps: representing entities and relations, defining a scoring function, and learning entity and relation representation~\cite{lin2015learning}. Translation-based embedding techniques such as TransD~\cite{TransD} and TransE~\cite{TransE} are the earliest embedding methods that represent the neighbourhood of nodes as well as the kind of relations that exist to the neighbouring nodes. 

Using translation-based embedding methods, embeddings are generated by treating relations as translations from the head entity to the tail entity~\cite{karim2019drug}. Embeddings are created such that $\mathbf{h}\bigoplus\mathbf{r} \approx \mathbf{t}$, by preserving a proximity measure defined on graph $\mathcal G$\footnote{\scriptsize{The $\bigoplus$ operator could be different w.r.t GE models.}}. The resultant embedding vectors for the entities and relations in in the KG are a denser and more efficient representation of the domain that can more easily be used for many downstream tasks. However, most translation-based embeddings are limited in their capacity to model complex and diverse objects, including important properties of relations, such as symmetric, transitive, one-many, many-to-one, and many-many relations in KGs~\cite{FTE}. Therefore, GE methods such as RDF2Vec\cite{RDF2Vec}, SimpleIE~\cite{SimpleIE}, KGloVe~\cite{cochez2017global}, and CrossE~\cite{CrossE} have been proposed. 

On the other hand, the GraphSAGE model learns embeddings of unlabeled graph nodes by leveraging the graph structure and features of the nodes. Being an inductive embedding technique, GraphSAGE allows extracting embeddings of unseen nodes, without the need to re-training. Unlike another embedding model such as Node2Vec which learns a look-up table of node embeddings instead of training individual embeddings for each node, GraphSAGE learns a function that generates embeddings by sampling and aggregating attributes from each node's local neighbourhood and combines those with the node's own attributes~\cite{GraphSAGE}. As real-life transaction network graphs are large-scale, some works were published with a view on scalability. In particular, FastGCN outperformed GCN and GraphSAGE in various benchmark datasets by as much as two orders of magnitude without sacrificing accuracy~\cite{weber2018scalable}. Weber et al.~\cite{weber2018scalable} trained GCN and FastGCN models on a large synthetic graph of 1 million nodes and 9 million edges. Although the GCN model significantly outperformed LR, the RF model turns out to be the best classifier, even outperforming the GCN. Another recent work~\cite{Mark2019} using a two-layer GCN model on a real transaction graph called \emph{Elliptic} from Bitcoin blockchain, shows that the GCN model outperforms linear models like LR, comparable to MLP, but underperforms in comparison with RF. They showed that the inclusion of non-local information -- specifically information about the neighbours of a centre node always improves the performance of each model. 

The majority of these models take into consideration only spatial information, thereby largely ignoring temporal information. However, transaction graphs are dynamic, where each transaction has an associated timestamp. To learn knowledge from such a dynamic graph, EvolveGCN~\cite{pareja2020evolvegcn} is proposed to extract node embeddings by coupling both spatial- and temporal information. EvolveGCN evolves along the temporal axis, where a recurrent neural network~(RNN) is used for evolving the GCN model's parameters. EvolveGCN is flexible for modelling temporal data since it is not reliant on node embeddings. Dynamic graph transformer~(DGT)~\cite{liu2021anomaly} is another approach in which the transformer model is composed of two modules: the \emph{transformer} and the \emph{pooling}. The transformer module captures the cross-domain knowledge by the attention mechanism, where the final attention layer generates the informative node embeddings. 

\section{Methods}\label{sec:methods}
We employ a semi-supervised graph learning technique on transaction graphs in order to identify nodes involved in potential money laundering. For this, we employ both {pipeline} and \emph{end-to-end} approaches. For the former, an embedding model is first trained to generate node embeddings that are used to train the ERT, GBT, and RF classifiers. For the latter, the node classification is performed in an end-to-end setting, without requiring training any separate classifiers. 

\begin{figure*}[h]
    \centering
    \includegraphics[width=\textwidth]{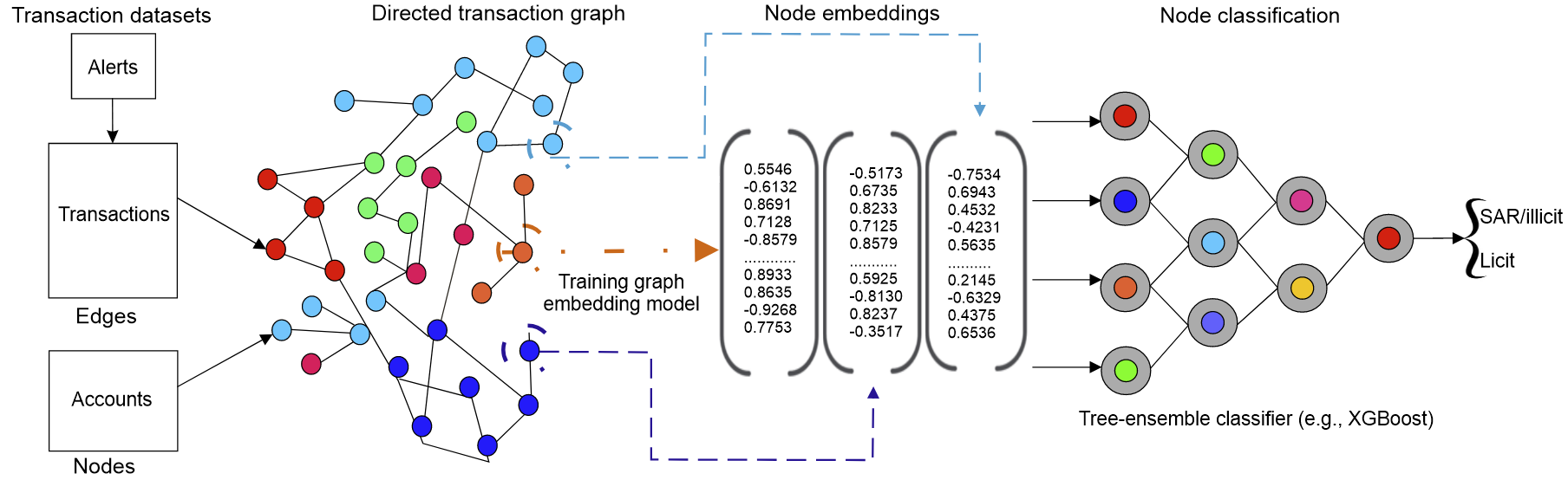}
    \caption{Workflow of our pipeline method for identifying nodes potentially involved in money laundering activities}
    \label{fig:workflow_of_proposed_approach}
\end{figure*}

\subsection{Problem formulation}
Let $\mathcal{G}=({V}, {E})$ be a money transfer or transaction graph, where nodes ${V}$ represent accounts and edges ${E}$ represent transfers. Let ${V}={X} \cup {W} \cup {Y}$, where ${W}$ is the inner accounts of the bank, and ${X}$ and ${Y}$ are sets of outer accounts. ${X}$ is the set of accounts that have the net transfer of money into the bank, and ${Y}$ is the set of accounts that have a net transfer out of the bank. An edge $(i,j) \in {E}$ indicates that account $v_i$ transfers money into another account $v_j$ for $v_i, v_j \in {V}$ and $e_{ij}$ is the amount of money transferred. Based on this setting, a directed KG can be represented as triplet facts ${(h, r, t) \in F}$ such that $\mathcal{G}=({V}, {E}, {F})$ to denote a link $r\in R$ between the head $\in V$ and the tail $t\in E$. % of the triple. 

Now using annotations from the alerts, the task in a semi-supervised learning setting is embedding the nodes into a lower dimensional vector space, followed by using the embedding vectors to train a binary classifier to predict the suspiciousness of a given target node in the graph via direct or indirect connections to nodes known to be suspicious. Let the embedding model $\Gamma$ embed each node of graph $\mathcal G$ into a lower dimensional vector space, yielding a set of vectors $\vec{V}$. More formally, given a graph $\mathcal G$, a node embedding is a mapping $\Gamma:{v_i} \rightarrow \vec{v_i} \in \mathbb{R}, \forall i \in [N]$ by capturing the information of the graph\footnote{\scriptsize{Depending on different embedding methods $\Gamma$ and embedding dimensions, different embeddings vector can be generated for the entities.}}, where $d$ is the dimension of the embeddings and $N$ is the number of nodes. The task is now to train a classifier $f$ on $\vec{V}$ in order to predict if a node is of suspiciousness, where the prediction $\hat{y_i}$ for embedding vector $\vec{v_i}$ for the $i^{th}$ node can be defined as follows: 

\begin{equation}
    \hat y_i=f(\vec{v_i}) = \left\{\begin{array}{ll}{1,} & {\text{if flagged, e.g., SAR or illicit}} \\ {0,} & {\text{otherwise.}}\end{array}\right.  
    \label{eq:ddi_}
\end{equation} 

For the sake of employing a semi-supervised learning paradigm, we remove a certain per cent of the nodes~(including edges from these nodes to any other nodes in the network), followed by training a GE model on the reduced sub-graph. During the inferencing, we generate the embeddings of the removed nodes using the trained GE model that are subsequently used to predict the labels of the nodes originally held out after re-inserting them in the network. 

\subsection{Generating directed graphs}\label{graph_generation_process}
We generate a directed transactional graph in five steps:

\begin{itemize}
  \setlength\itemsep{0.3em}
    \item \textbf{Step-1:} Load transactions, alerts\footnote{\scriptsize{e.g., containing falgs like SARs or illicitness.}}, and party datasets.
    \item \textbf{Step-2:} Define user defined utility functions for preprocessing and encoding categorical features.
    \item \textbf{Step-3:} Generate graph nodes and edges.
    \item \textbf{Step-4:} Form a directed graph for nodes and edges.
    \item \textbf{Step-5}: Annotate the nodes with alerts datasets or additional features\footnote{\scriptsize{i.e., labelled SARs indicating if a node was part of a previously known money laundering scheme or illicit.}}. 
\end{itemize}

\begin{figure*}
	\centering
	\begin{subfigure}{0.48\linewidth}
		\centering
		\includegraphics[width=0.75\textwidth]{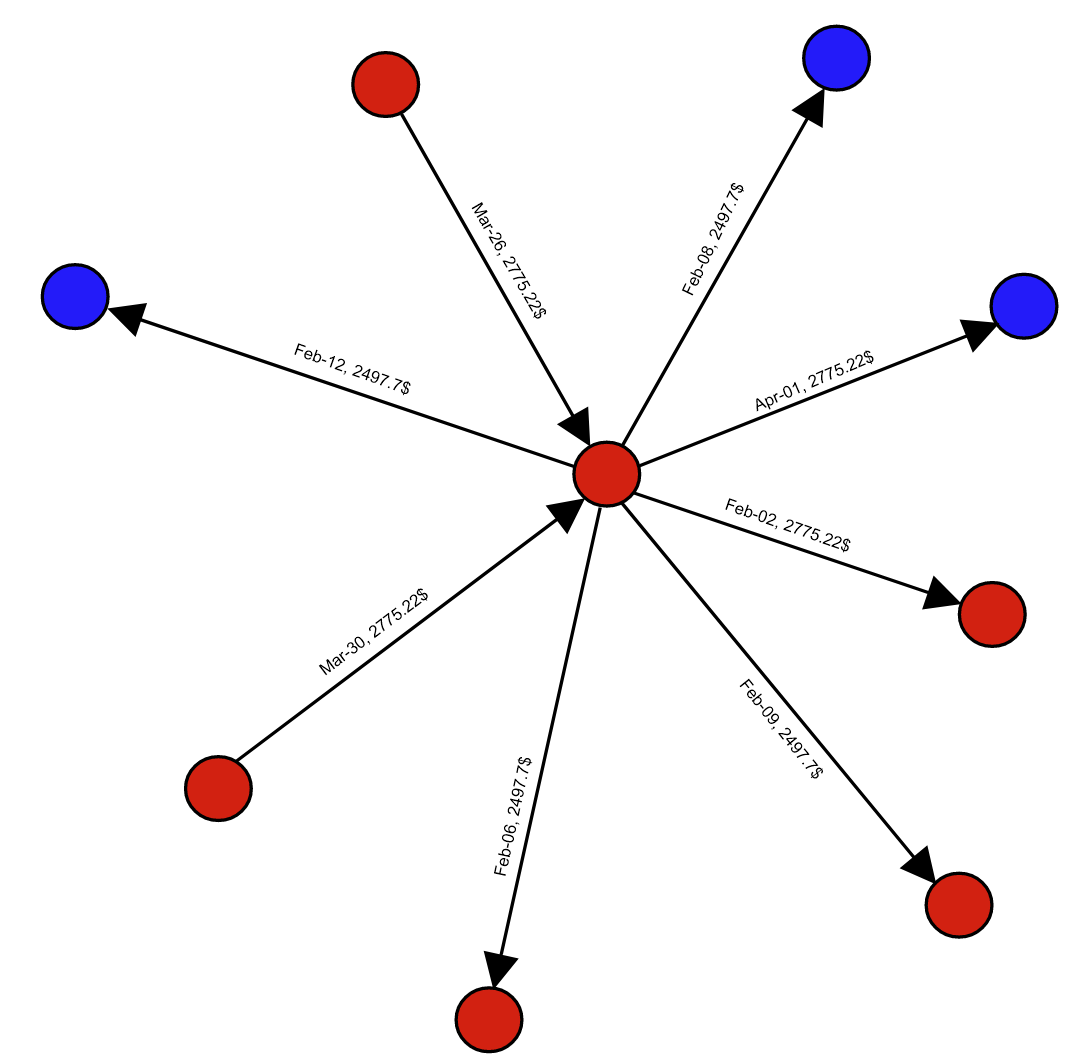}
		\caption{Gather scatter}
        \label{fig:gather_scatter}
	\end{subfigure}\hfill
	\begin{subfigure}{0.48\linewidth}
		\centering
		\includegraphics[width=0.75\textwidth]{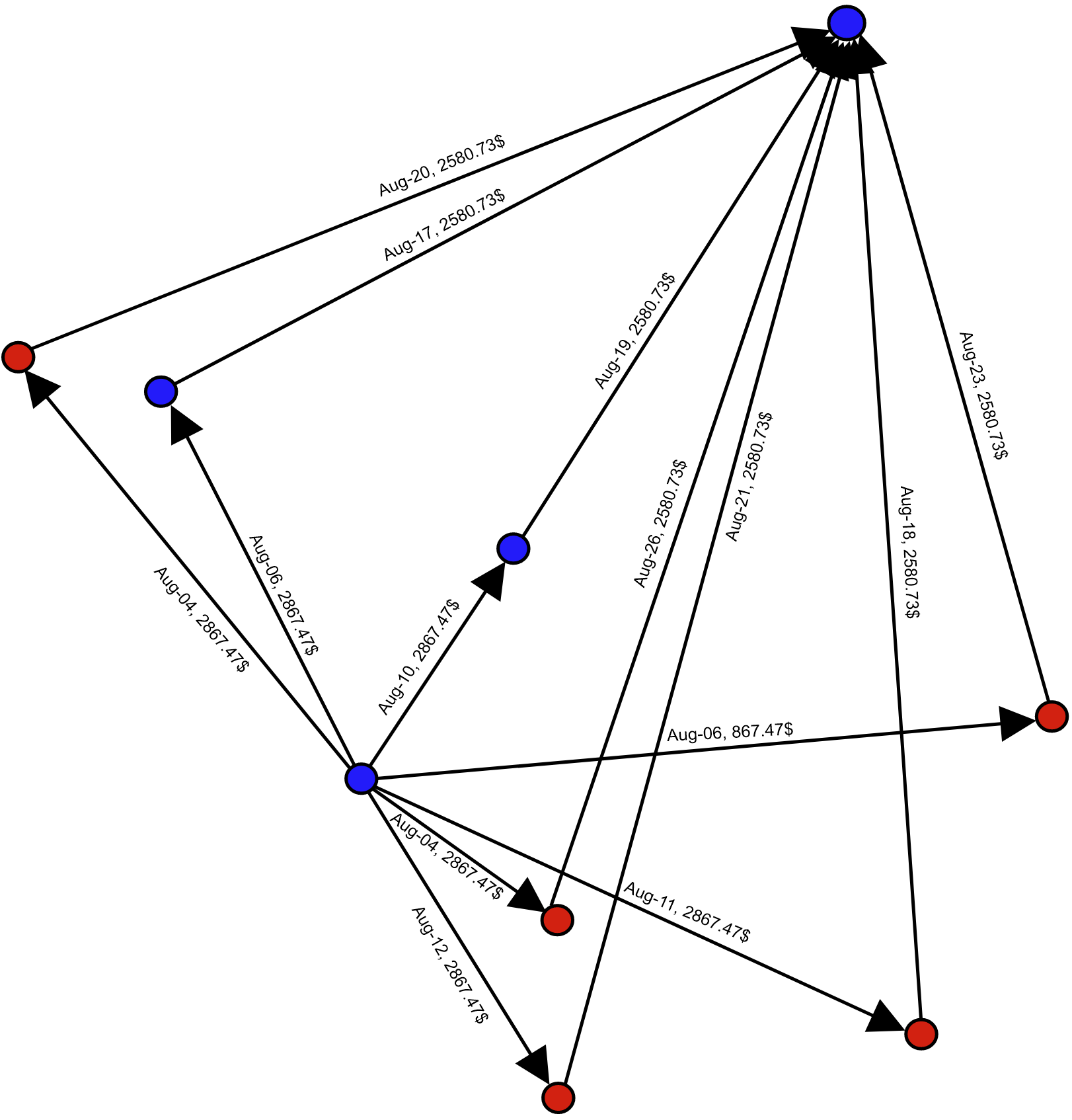}
		\caption{Scatter gather}
        \label{fig:scatter_gather}
	\end{subfigure}
	\begin{subfigure}{0.48\linewidth}
		\centering
		\includegraphics[width=0.75\textwidth]{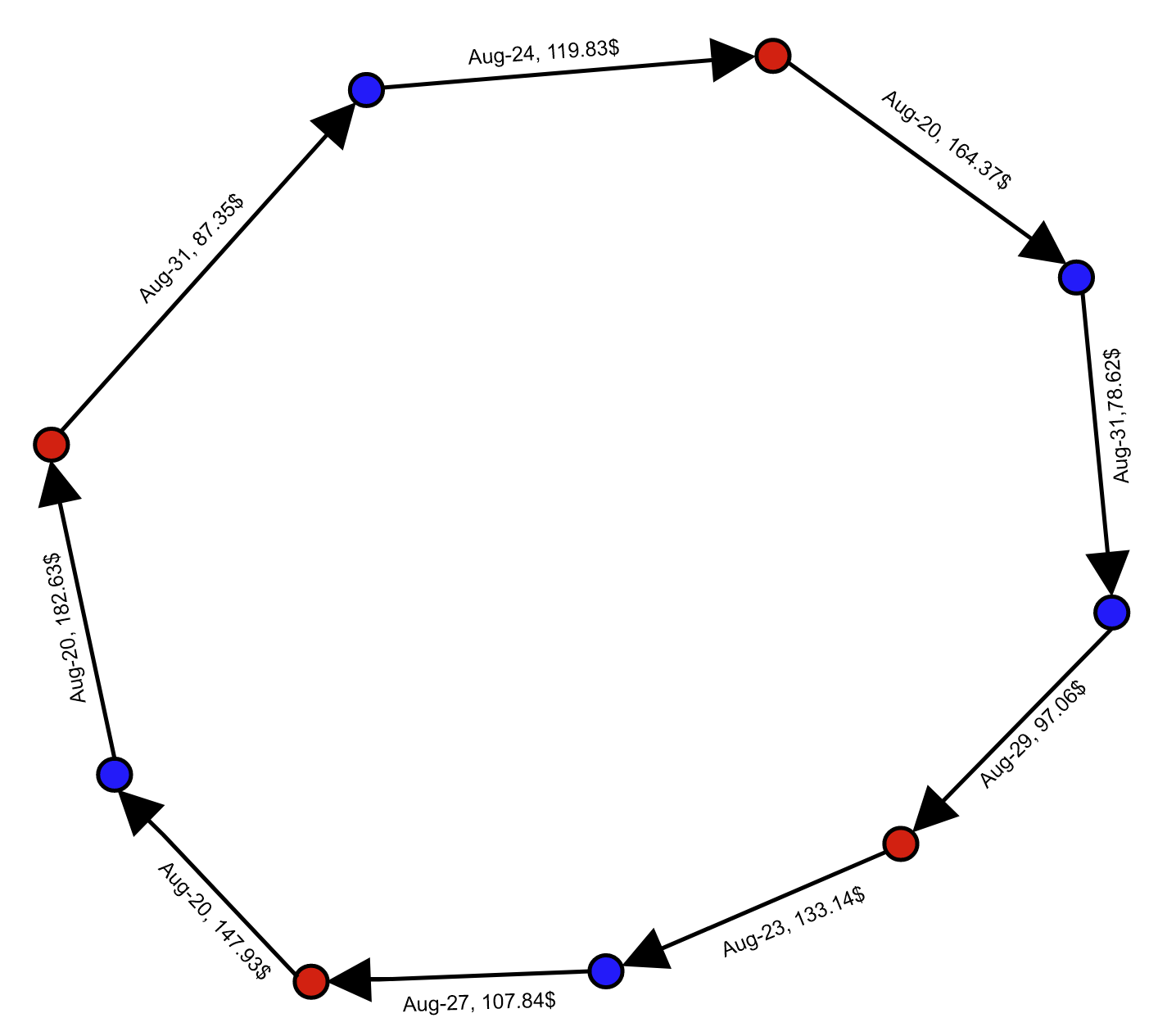}
		\caption{Cycle}
        \label{fig:cycle}
	\end{subfigure}
	\caption{Different (alter) types of transactions in the graph} 
	\label{fig:transaction_types}
\end{figure*}

Examples of different types of transactions that are flagged with ``alert types'' are shown in \cref{fig:transaction_types}. 

\subsection{Graph embeddings}
Since ML classifiers do typically expect their input as fixed-length vectors, we employ different unsupervised graph representation learning techniques to generate node embeddings using the Word2vec, Node2vec, Attri2Vec, GraphSAGE, and DGT models that represent the neighbourhood of a node and their relations to the neighbouring nodes. 

Using \emph{Word2vec} and \emph{Node2Vec}, a corpus of text $\mathcal{C}$ is generated by performing uniform random walks starting from each entity in the graph~\cite{cochez2017biased}. Then, $\mathcal{C}$ of edge-labelled random walks are used as the input for learning embeddings of each node using the skip-gram~(SG) Word2vec~\cite{mikolov2013efficient} model. From a given a sequence of facts $(w_1,w_2,...,w_n)\in \mathcal{C}$, the SG model aims to maximize the average log probability $L_p$ according to the context within the fixed-size window~\cite{mikolov2013efficient}: 

\begin{equation}
    L_p= \frac{1}{N} \sum_{n=1}^{N} \sum_{-c \leq j \leq c, j \neq 0} \log p\left(w_{n+j} | w_{n}\right),
    \label{eq:log}
\end{equation}

where $c$ represents a context. To define $p\left(w_{n+j} | w_{n}\right)$, we use negative sampling by replacing $\log p\left(w_{O} | w_{I}\right)$ with a function to discriminate target words $(w_o)$ from a noise distribution $P_n(w)$ drawing $k$ words from $P_n(w)$~\cite{karim2019drug}:

\begin{equation}
    \log \sigma\left(v_{w_{O}}^{\prime \top} v_{w_{I}}\right)+\sum_{i=1}^{k} \mathbb{E}_{w_{i}} \sim_{P_{n}(w)}\left[\log \sigma\left(-v_{w_{i}}^{\prime \top} v_{w_{I}}\right)\right].
    \label{eq:word2vec}
\end{equation}

The embedding of a concept $c$ occurring in corpus $\mathcal{C}$ is the vector $\vec v_s$ in \cref{eq:word2vec} derived by maximizing \cref{eq:log}. Technically both Word2vec and Node2Vec algorithms follow is a 2-step representation learning algorithm: i) using second-order random walks to generate sentences from a graph\footnote{\scriptsize{A sentence is a list of node ids. A corpus is the set of all sentences.}}, ii) the corpus is then used to learn an embedding vector for each node in the graph. Each node id is considered a unique word/token in a dictionary that has a size equal to the number of nodes $N$ in our graph $\mathcal{G}$. 

The Attri2Vec model~\cite{attri2vec} is trained to learn node representations by performing linear/non-linear mapping on node content attributes. To make the learned node representations respect structural similarity, DeepWalk/Node2Vec learning mechanism is employed to make nodes sharing similar random walk context nodes represented closely in the subspace. For each (target, context) node pair $\left(v_i, v_j\right)$ from random walks, Attri2Vec learns the representation $\vec v_i$ for the target node $v_i$ by using it to predict the existence of context node $v_j$ based on a three-layer neural network~\cite{attri2vec}. The representation of a node $\vec {v_i}$ in the hidden layer is then obtained by multiplying its raw content feature vector in the input layer with the input-to-hidden weight matrix $W_{in}$, followed by an activation function~\cite{attri2vec}. 

For a given large set of ``positive'' (target, context) node pairs generated from random walks and an equally large set of ``negative'' node pairs that are randomly selected from graph $\mathcal{G}$ according to a certain distribution, GraphSAGE learns a binary classifier that predicts whether arbitrary node pairs are likely to co-occur in a random walk performed on the graph~\cite{GraphSAGE}.  The GE models we train so far take into consideration only spatial/structural information, thereby not considering temporal information. Therefore, we learn knowledge from such a dynamic graph, with the hypothesis that the anti-money laundering could be benefited from it since DGT is able to couple both spatial- and temporal information capturing simultaneously~\cite{liu2021anomaly}. 

Taking into consideration the dynamic nature of the transaction graph, let node $v_1^t$ and $v_2^t$ be involved in a transfer at time $t$, where their common connections have had several transactions in the previous timestamps. Then, this temporal relation can be represented as $u_1^{t-1}-u_2^{t-1}$ and $u_1^{t-2}-u_2^{t-2}$~\cite{liu2021anomaly}. To extract spatial-temporal knowledge, encodings of the nodes are aggregated within a substructure node set into node embeddings. Attention layers are utilized to exchange the information of different nodes, where a single attention layer is represented as follows~\cite{liu2021anomaly}:

\begin{equation}
    \mathbf{H}^{(l)}=\operatorname{att}\left(\mathbf{H}^{(l-1)}\right)=\operatorname{softmax}\left(\frac{\mathbf{Q}^{(l)} \mathbf{K}^{(l) \top}}{\sqrt{d}}\right) \mathbf{V}^{(l)},
    \label{eq:aggregating}
\end{equation}

where $\mathbf{H}^{(l)}$ and $\mathbf{H}^{(l-1)}$ is the output embedding for the $l$ and $(l-1)^{th}$ layer, respectively; $d$ is the dimension of node embedding, $att$ signifies the self-attention operation; $\mathbf{Q}^{(l)}, \mathbf{K}^{(l)}, \mathbf{V}^{(l)} \in \mathbb{R}^{(\tau(k+2)) \times d}$ are the query-, key-, and value matrices for feature transformation and information exchange, respectively that can be represented as~\cite{liu2021anomaly}:

\begin{equation}
    \left\{\begin{array}{l}
    \mathbf{Q}^{(l)}=\mathbf{H}^{(l-1)} \mathbf{W}_Q^{(l)}, \\
    \mathbf{K}^{(l)}=\mathbf{H}^{(l-1)} \mathbf{W}_K^{(l)}, \\
    \mathbf{V}^{(l)}=\mathbf{H}^{(l-1)} \mathbf{W}_V^{(l)},
    \end{array}\right.
\end{equation}

where $\mathbf{W}_Q^{(l)}, \mathbf{W}_K^{(l)}, \mathbf{W}_V^{(l)} \in \mathbb{R}^{d \times d}$ are the learnable parameter matrices of the $l$-th attention layer. In an attention layer, $\mathbf{Q}^{(l)}$ and $\mathbf{K}^{(l)}$ calculate the contributions of different nodes' embeddings, while $\mathbf{V}^{(l)}$ projects the input into a new feature space that is combined as of \cref{eq:aggregating} to acquire the output embedding of each node by aggregating the information of all nodes adaptively~\cite{liu2021anomaly}.

The input of the transformer module $\mathbf{H}^{(0)}$ represents the encoding matrix of the target edge $\mathbf{X}\left(e_{\mathrm{tgt}}^t\right)$ by setting $d=d_{{enc}}$ to align the dimension. The output of the final attention layer $\mathbf{H}^{(L)}$ is extracted as the output node embedding matrix $\mathbf{\vec Z}$ of the transformer module, where each row represents an embedding vector of a node~\cite{liu2021anomaly}. 

\subsection{Training of classifiers}
We train decision trees~(DTs) and their ensemble models such as RF, extremely randomized trees~(ERT), and gradient-boosted trees~(GBT) on learned embeddings. DTs exploit tree structures, where internal nodes represent feature values w.r.t boolean conditions and leaf nodes represent predicted labels. DT iteratively splits $X^{*}$\footnote{Let it represents the set of embedding vectors $\vec V$.} into multiple subsets w.r.t to threshold values of features at each node until each subset contains instances from one class only. Each branch in a DT represents a possible outcome, where the relationship between prediction $\hat{y}^{*}_{i}$ and feature $\vec x^{*}_i$ can be defined as~\cite{di2019surrogate}: 

\begin{align}
    \hat{y}^{*}_{i}=f\left({x}^*_{i}\right)=\sum_{j=1}^{N} c_{j} I\left\{{X}^{*}_{i} \in R_{j}\right\},
\end{align}

where each sample $x^{*}$ reaches exactly one leaf node, $R_{j}$ is the subset of the data representing the combination of rules at each internal node, and $I\{.\}$ is an identity function~\cite{di2019surrogate}. In the case of tree ensemble models, the prediction function $f(\mathrm{x^*})$ is defined as the sum of individual feature contributions plus the average contribution for the initial node in a DT for the dataset and $K$ possible class labels that change along the prediction path after every split, along with the information which a feature caused the split~\cite{al2021cdrgi}:

\begin{align}
    f(x)=c_{full}+\sum_{k=1}^{M} \sigma(x, k),
\end{align}
where $c_{full}$ is the average of the entire training set $X$ dataset~(initial node), $\mathrm{M}$ is the total number of features.

For graph-based node classification, which is technically predicting the label of a node $u$ at time $t$, we follow the usual training procedures for EvolveGCN~\cite{pareja2020evolvegcn} and FastGCN~\cite{Chen2018}, followed by the standard GCN-like approach: the activation function of the last graph convolution layer is set to sigmoid so that $h_t^u$ is a probability vector over two probable classes. 

\section{Experiments}\label{sec:exp}
In this section, we report and analyse experiment results. 

\subsection{Datasets}
We tested approach on two datasets: \emph{AMLSim} and \emph{Elliptic dataset}. The AMLSim is a multi-agent simulation platform tailored for an AML problem, where each agent behaves as a bank account transferring money to other agent accounts, and where a small number of agents conduct nefarious activity modelled on real-world patterns. we generate a dynamic directed transaction graph containing semi-realistic suspicious activities, based on the following information and graph generation process described in \cref{graph_generation_process}: 

\begin{itemize}
  \setlength\itemsep{0.3em}

    \item \textbf{Accounts}: contains the information about all the bank accounts whose transactions are monitored.
    \item \textbf{Alerts}: contains the list of transactions that triggered an alert according to AML guidelines.
    \item \textbf{Transactions}: contains the list of all the transactions with information about sender and receiver accounts.
\end{itemize}

Each node in the graph represents an account with attributes such as account number, account type, owner name, and date/time created. Nodes are designated with \emph{cash in}, \emph{`cash out}, \emph{debit}, \emph{payment}, \emph{transfer} or \emph{deposit} activities and are of either organization or individual types. Each edge has a transaction ID, amount, and time stamp. The data is sparsely labelled with flagged transactions\footnote{\scriptsize{Violating both volume and velocity rules.}} and SARs\footnote{\scriptsize{Confirmed suspiciousness.}}.

The Elliptic dataset~\footnote{\scriptsize{\url{https://www.kaggle.com/datasets/ellipticco/elliptic-data-set}}} is a graph network of Bitcoin transactions with handcrafted features. All features are constructed using only publicly available information. This anonymized data set is a transaction graph collected from the Bitcoin blockchain. The Elliptic Dataset maps Bitcoin transactions to real entities in two categories~\cite{Mark2019}:

\begin{itemize}
  \setlength\itemsep{0.3em}
    \item \textbf{Licit}: Exchanges, wallet providers, miners, licit services.
    \item \textbf{Illicit}: Scams, malware, terrorist organizations, ransomware, Ponzi schemes, etc.
\end{itemize}

The graph contains 203,769 node transactions and 234,355 directed edge payments flow out of which 2\% are illicit, 21\% are licit. The remaining 77\% samples are labelled as `unknown' transactions. 
Each node has 166 features associated: the first 94 features represent local information\footnote{\scriptsize{Timestep, number of inputs/outputs, transaction fee, output volume and aggregated figures, e.g., average BTC received/spent by inputs/outputs, the average number of incoming/outgoing transactions.}} The remaining 72 features represent aggregated features that are obtained using transaction information one-hop backwards/forwards from the centre node\footnote{\scriptsize{Max/min standard deviation and correlation coefficients of neighbour transactions w.r.t number of inputs/outputs, transaction fee, etc.}}. Time steps are associated with each node, representing an estimated time when the transaction is confirmed. There are 49 distinct timesteps evenly spaced with an interval of 2 weeks. 

\subsection{Experiment settings}
We use open source StellarGraph\footnote{\scriptsize{\url{https://github.com/stellargraph/stellargraph}}} library to compute node embeddings and are based on BiasesRandomWalk and Word2Vec from the gensim library. The DGT model is trained based on its PyTorch implementation`\footnote{\scriptsize{\url{https://github.com/yuetan031/TADDY_pytorch}}}. The RF and ERT models are based on scikit-learn's ensemble methods. As of GBT, we train its XGBoost implementation\footnote{\scriptsize{\url{https://xgboost.readthedocs.io/}}}. 
While RDF2Vec was trained using a skip-gram model by setting a window size of 5 with graph walk at depth 5 and 500 walks per entity. To observe the efforts of embedding dimension, for each embedding model, the value of $d$ is selected from $\{32, 64, 128, 256, 300\}$. As for the DGT model, number of layers $L$ is selected from $\{1, 2, 3, 4, 5\}$. 

As for the pipeline approach, Node2vec, Attri2Vec, GraphSAGE, and DGT models are first trained to generate node embeddings that are then used to train classifiers ERT, GBT, and RF. As of end-to-end approach, we train GCN~\cite{Kipf2017} with batched training, FastGCN~\cite{Chen2018} that reduces the training cost through neighbourhood sampling, and EvolveGCN
to capture the dynamism by evolving GCN parameters. Both models were configured to have 128 hidden units and trained with AdaGrad with varying learning rates and batch sizes. % to optimize the \texttt{binary cross-entropy} loss. 
Further, since classes are imbalanced in the case of both datasets, we trained GCN, FastGCN, and EvolveGCN models using a \emph{weighted cross-entropy loss} to provide higher importance to minority class~(i.e., illicit/SARs samples). 

For each experiment, we remove 10 to 20\% of the nodes, followed by training the GE models on the reduced sub-graph. During the inferencing, we generate the embeddings of the removed nodes using the trained GE model that are subsequently used to predict the labels of the nodes originally held out after re-inserting them in the network. Thus, for each experiment, 80\% of the data is used for the training and validating in which the best hyperparameters were produced via random search and 5-fold cross-validation. We used the area under the precision-recall curve~(AUPR), and Matthias correlation coefficient~(\texttt{MCC}) along with the AUC and F1-scores to measure the performance of each classifier. 

\subsection{Analysis of node classifications} 
\Cref{table:result_pipeline_amlsim} and \cref{table:result_pipeline_bitcoin} summarizes the results of the prediction task based on pipeline methods. As seen, the \emph{DGT+XGBoost} combination outperformed all other combinations, covering both datasets. The ROC curves show that all the classifiers perform worse when trained on embeddings generated by the Node2Vec and Attri2Vec models. The performance of GraphSAGE+XGBoost is comparable to DGT+XGBoost. 

However, a general observation is that the end-to-end methods outperformed each and every pipeline methods. In particular, EvolveGCN outperforms all the pipeline methods with tree-ensemble classifiers, indicating the effectiveness of end-to-end methods compared over pipeline methods. Further, EvolveGCN consistently outperforms both Skip-GCN and FastGCN, although the improvement is not very substantial for both datasets. 

\begin{table}
	\caption{Pipeline methods on AMLSim}
	\label{table:result_pipeline_amlsim}
 \centering
 \vspace{-3mm}
	\scriptsize{
	\begin{tabular}{c|c|c|c|c}
		 \hline
		\textbf{Emb. model} & \textbf{Classifier} & \textbf{AUPR} & \textbf{F1-score} & \textbf{MCC}\\ \hline
		\multirow{3}{*}{Node2Vec} 
		& ERT & 0.746 & 0.753 & 0.643\\
		& RF & 0.751 & 0.760 & 0.651\\
		& XGBoost  & 0.806 & 0.801 & 0.752 \\ \hline
		\multirow{3}{*}{Attri2Vec} 
		& ERT & 0.762 & 0.783 & 0.651\\
		& RF & 0.775 & 0.782 & 0.669\\
		& XGBoost  & 0.815 & 0.810 & 0.7692 \\ \hline
		\multirow{3}{*}{GraphSAGE} 
		& ERT & 0.786 & 0.801 & 0.679\\
		& RF & 0.802 & 0.804 & 0.675\\
		& XGBoost  & 0.815 & 0.816 & 0.701 \\ \hline
		\multirow{3}{*}{\textbf{DGT}} 
		& ERT & 0.797 & 0.803 & 0.671\\
		& RF & 0.813 & 0.825 & 0.693\\
		& \textbf{XGBoost}  & \textbf{0.833} & \textbf{0.832} & \textbf{0.715} \\ \hline
	\end{tabular}}
\end{table}

\begin{table}
	\caption{End-to-end methods on AMLSim}
	\label{table:result_e2e_amlsim}
 \centering
 \vspace{-3mm}
	\scriptsize{
	%\centering
	\begin{tabular}{c|c|c|c}
		\hline
		\textbf{Model} & \textbf{AUPR} & \textbf{F1-score} & \textbf{MCC}\\ \hline
		Skip-GCN & 0.834~(0.792) &  0.915~(0.875) & 0.881~(0.763)\\ \hline 
		FastGCN & 0.841~(0.804) &  0.927~(0.890) & 0.903~(0.781)\\ \hline
		\textbf{EvolveGCN}  & \textbf{0.869~(0.813)} & \textbf{0.934~(0.902)} & \textbf{0.891~(0.773)} \\ \hline
	\end{tabular}}
\end{table}

\begin{figure*}
	\centering
	\begin{subfigure}{0.48\linewidth}
		\centering
		\includegraphics[width=90mm,height=70mm]{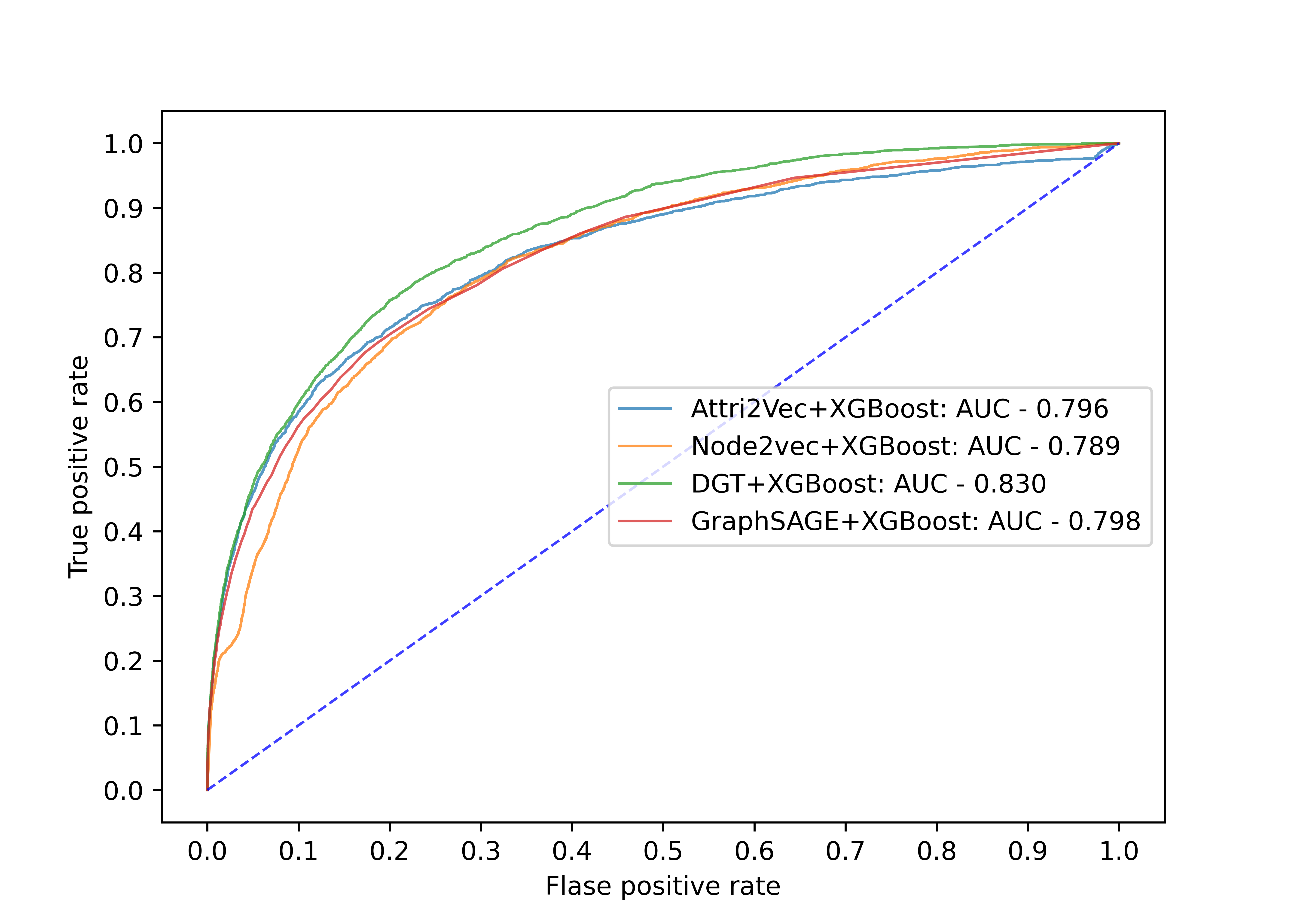}
		\caption{Pipeline methods}
        \label{fig:auc_roc_pipeline_amlsim}
	\end{subfigure} %\hfill
	\begin{subfigure}{0.48\linewidth}
		\centering
		\includegraphics[width=90mm,height=70mm]{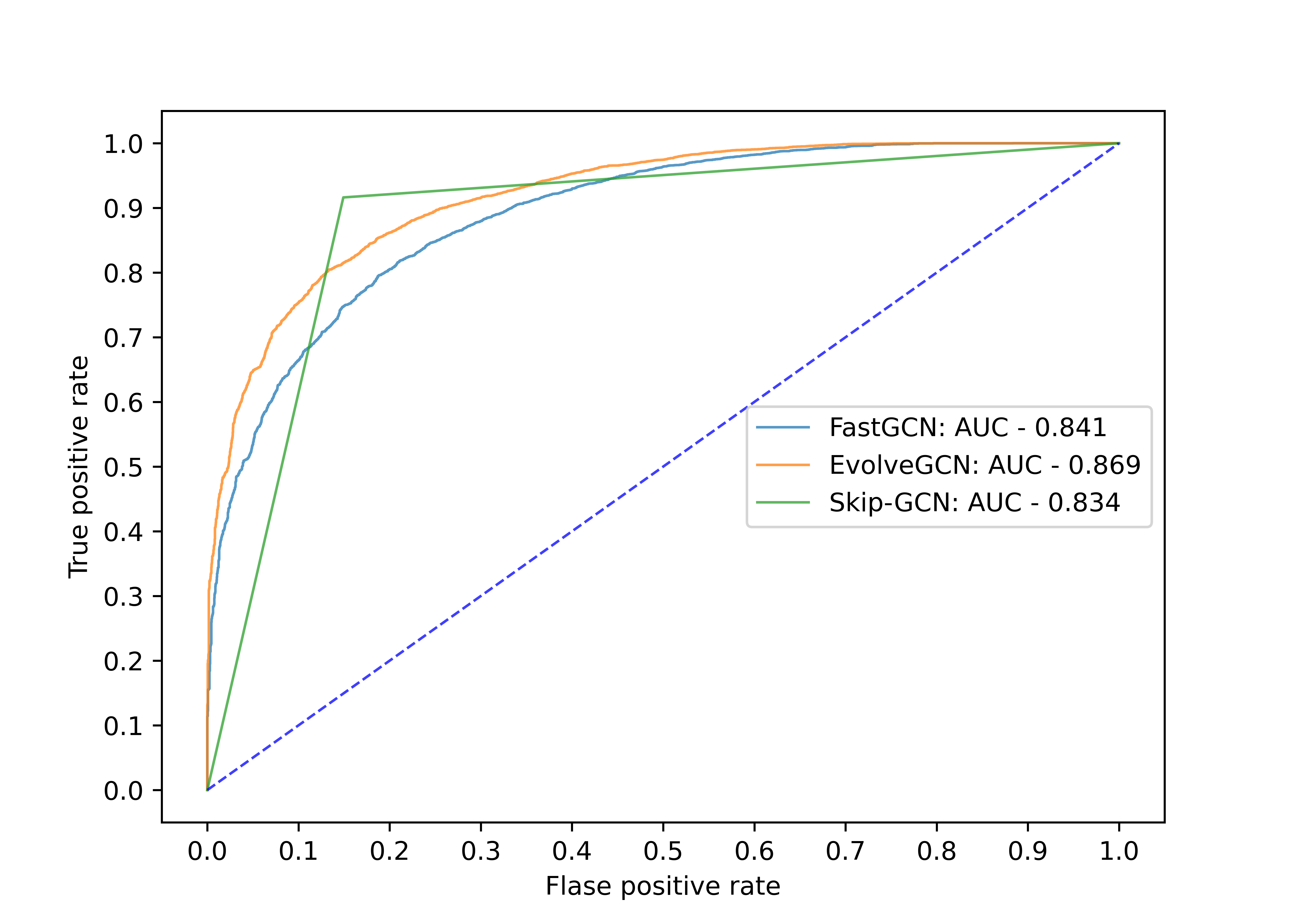}
		\caption{End-to-end methods}
        \label{fig:auc_roc_e2e_amlsim}
	\end{subfigure}
	\caption{Sensitivity vs. specificity for AMLSim dataset} 
	\label{fig:auc_roc_amlsim}
\end{figure*}

\begin{figure*}
	\centering
	\begin{subfigure}{0.49\linewidth}
		\centering
		\includegraphics[width=90mm,height=70mm]{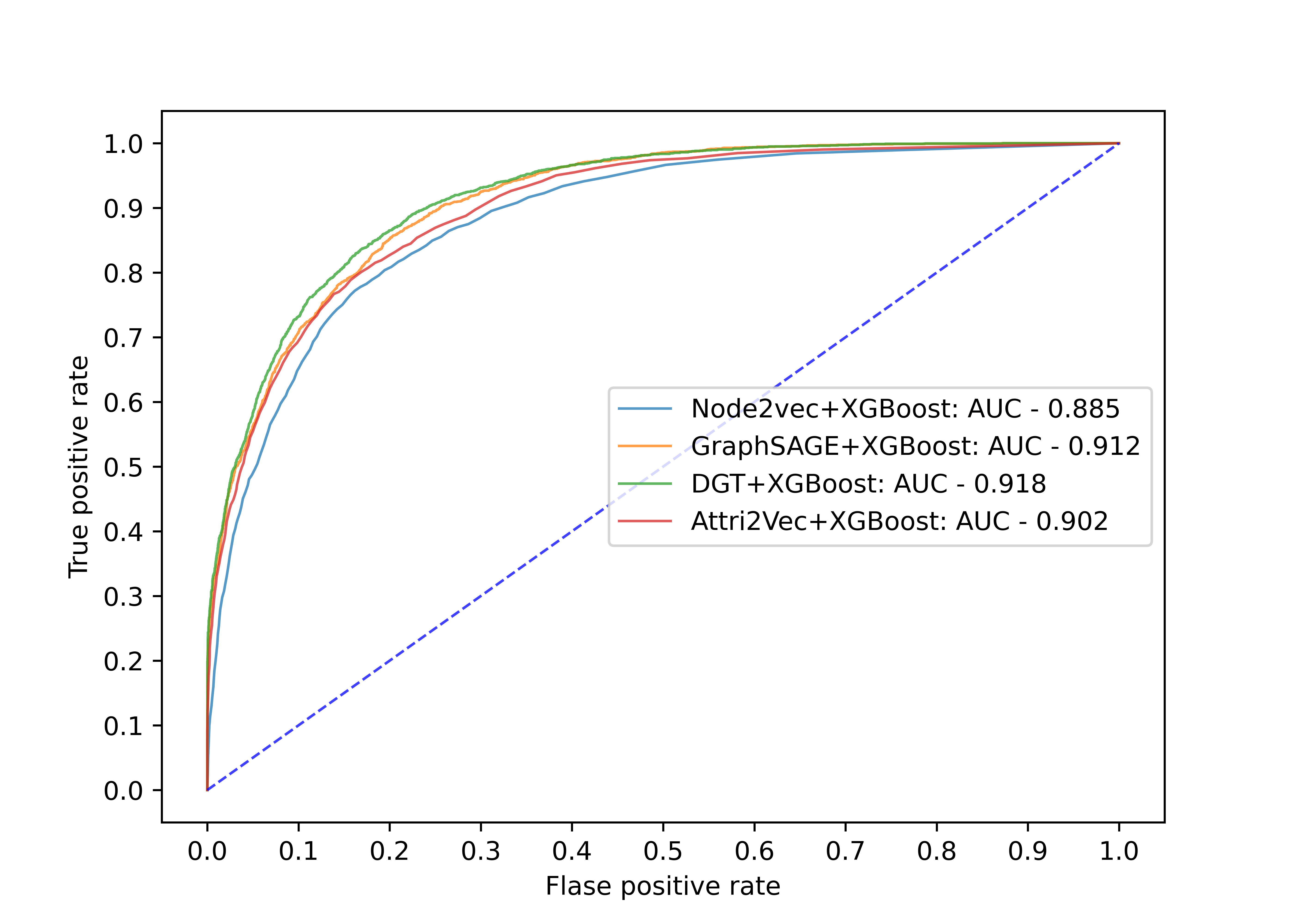}
		\caption{Pipeline methods}
        \label{fig:auc_roc_pipeline_elliptic}
	\end{subfigure} %\hfill
	\begin{subfigure}{0.49\linewidth}
		\centering
		\includegraphics[width=90mm,height=70mm]{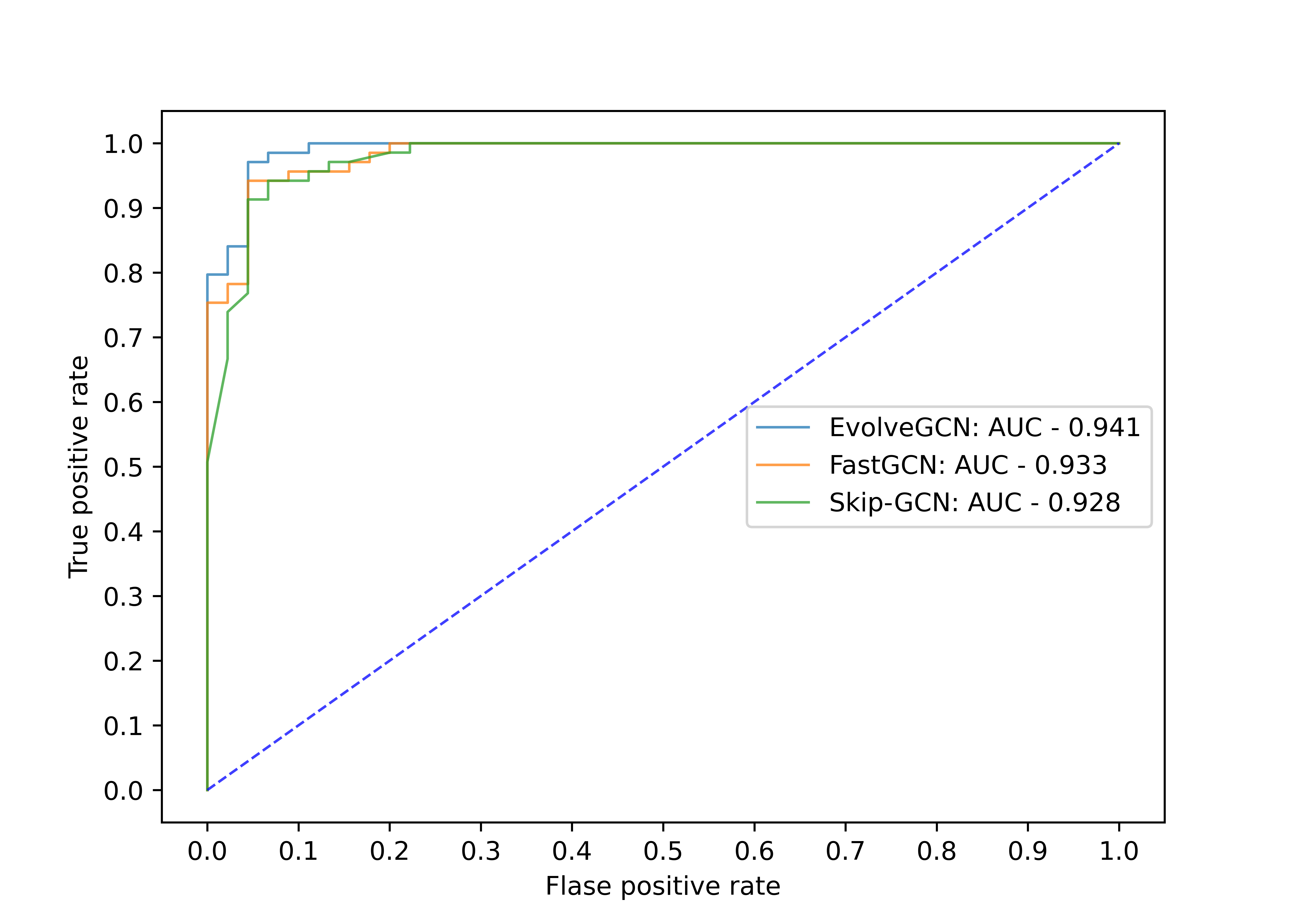}
		\caption{End-to-end methods}
        \label{fig:auc_roc_e2e_elliptic}
	\end{subfigure}
	\caption{Sensitivity vs. specificity for Elliptic dataset} 
	\label{fig:auc_roc_elliptic}
\end{figure*}

\begin{table}
	\caption{Pipeline methods on Elliptic}
	\label{table:result_pipeline_bitcoin}
 \centering
 \vspace{-3mm}
	\scriptsize{
	\begin{tabular}{c|c|c|c|c}
		 \hline
		\textbf{Emb. model} & \textbf{Classifier} & \textbf{AUPR} & \textbf{F1-score} & \textbf{MCC}\\ \hline
		\multirow{3}{*}{DGT} 
		& ERT & 0.885 & 0.854 & 0.726\\
		& RF & 0.907 & 0.897 & 0.753\\
		& XGBoost  & 0.918 & 0.915 & 0.792 \\  \hline
		\multirow{3}{*}{GraphSAGE} 
		& ERT & 0.874 & 0.865 & 0.743\\
		& RF & 0.891 & 0.882 & 0.778\\
		& XGBoost  & 0.912 & 0.905 & 0.782 \\  \hline
		\multirow{3}{*}{Attri2Vec} 
		& ERT & 0.815 & 0.824 & 0.653\\
		& RF & 0.821 & 0.832 & 0.665\\
		& XGBoost  & 0.902 & 0.894 & 0.673 \\  \hline
		\multirow{3}{*}{Node2Vec} 
		& ERT & 0.792 & 0.805 & 0.617\\
		& RF & 0.806 & 0.817 & 0.622\\
		& XGBoost  & 0.885 & 0.874 & 0.662 \\ \hline
	\end{tabular}
	}
\end{table}

\begin{table}
	\caption{End-to-end methods on Elliptic}
	\label{table:result_e2e_bitcoin}
 \centering
 \vspace{-3mm}
	\scriptsize{
	%\centering
	\begin{tabular}{c|c|c|c}
		\hline
		\textbf{Model} & \textbf{AUPR} & \textbf{F1-score} & \textbf{MCC}\\ \hline
		Skip-GCN & 0.928~(0.793) &  0.916~(0.873) & 0.854~(0.763)\\ \hline 
		FastGCN & 0.933~(0.805) &  0.925~(0.881) & 0.875~(0.781)\\ \hline
		\textbf{EvolveGCN}  & \textbf{0.941~(0.813)} & \textbf{0.934~(0.891)} & \textbf{0.891~(0.773)} \\ \hline
	\end{tabular}}
\end{table}

\subsection{Effects of temporal information} 
As for the AMLSim dataset, the XGBoost model mostly benefited from the temporal information captured by the DGT embedding model. Subsequently, this pipeline method outperformed other methods of this class. This is also the case for the Elliptic dataset, where same method outperformed GraphSAGE+XGboost. These results indicate that the dynamic models are more effective. A potential reason could be that the input features are already quite informative and on top of that GNNs or transformer were able to extract more abstract features that are relevant in distinguishing licit and illicit nodes. Overall, the representation learning capability of these models from input features is clearly reflected in the classification results.

\subsection{Effects of nodes embeddings + feature space} 
\begin{figure}[h]
    \centering
    \includegraphics[width=0.3\textwidth]{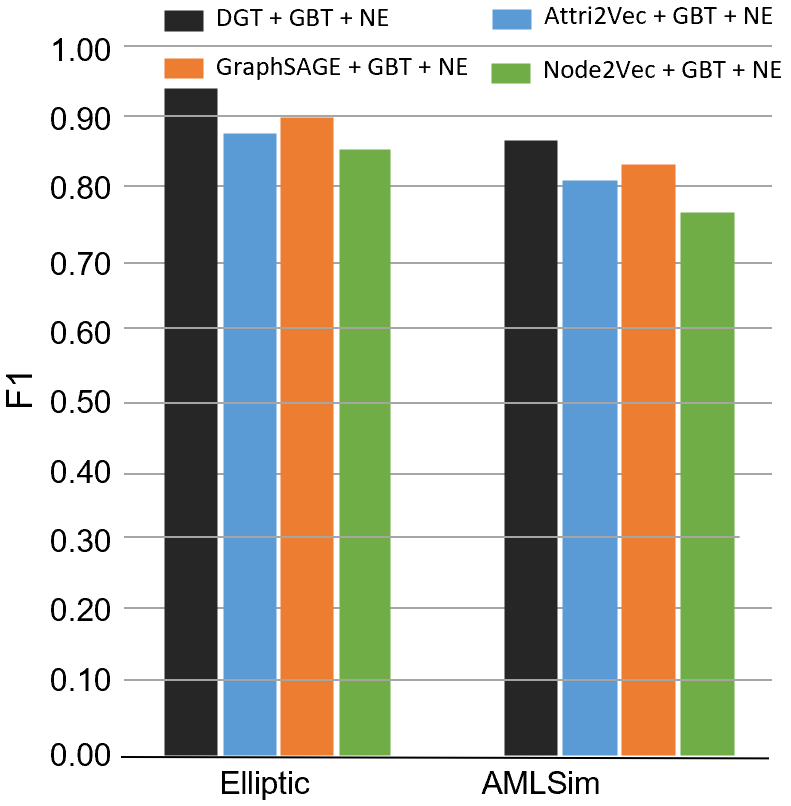}
    \caption{Effects of combining node embeddings with features}
    \label{fig:node_embedding_plus_feature_space}
\end{figure}

A recent approach~\cite{pareja2020evolvegcn} showed that aggregated information may lead higher F1 scores for anomaly detection like tasks. Inspired by this, we extended the feature space by combining node embeddings obtained from each embedding models and retrained the models. As shown in \cref{fig:node_embedding_plus_feature_space}, the node classification accuracy is slightly improves, by taking the global graph structure in this context into consideration. This somewhat becomes comparable to end-to-end methods. The micro averages for all pipeline approach is higher than 0.93. However, they are not very informative for highly imbalanced datasets. In the case of financial crime forensics, the minority illicit class is of main interest. Therefore, we plot the minority F1 scores for both datasets. 

\section{Conclusion and Outlook}\label{sec:con}
In this paper, we employed semi-supervised graph learning techniques on graphs of financial transactions in order to identify nodes involved in potential money laundering. We use SARs annotations from alerts as the ground truths. We trained different GE models such as Word2vec, Node2vec, Attri2Vec, GraphSAGE, and DGT models that embed the nodes into a lower dimensional vector space. Then, using the embeddings, we trained RF, GBT, and ERT classifiers to predict the suspiciousness of a given target node in the graph via direct or indirect connections to nodes known to be suspicious. Further, we critically reviewed existing AML methods and outlined their potential limitations. 

With the rapid development of a cashless society engaged in global economic exchange, the advent of cryptocurrency has catalyzed a paradigm shift in peer-to-peer transactions and extranational financial governance. Cryptocurrencies not only impose great challenges to AML but also increase difficulty across cryptocurrency types. Another challenge lies involving temporal dynamics with the emergence/disappearance of new entities in the blockchain. For example, Weber et al.~\cite{weber2018scalable} have shown that at time step the market may appear to follow \emph{Dark Market shutdown}. In such a situation, no models~(including EvolveGCN or DGT) would be able to capture such high volatility and consequently may not perform well. 

AML detection mechanisms as explored in this study are not immune to attacks, while deep models may lack robustness under adversarial attacks. A transactional dataset itself might be viewed as private since the transactional dataset contains sensitive personal information and data with a high monetary value for the organisation possessing it. Thus, any form of privacy attack that could potentially disclose privacy-sensitive information could be very costly. Moreover, an adversary might craft\footnote{\scriptsize{e.g., by adding noise or minor perturbations in the embedding space.}} a series of specific transactions to fool the AML detection system to classify an illicit transaction to be licit. Such adversarial attacks are real security threats for any financial systems~\cite{berman2019survey}. On the other hand, since the computed embeddings might need to be shared, which an attacker could apply techniques to reverse\footnote{\scriptsize{e.g., by employing a sophisticated approach for input reconstruct.}} to original input or link back to a natural person. Further, a deployed AML detection model could be directly attacked too. In particular, membership inference attacks~\cite{hu2022membership} can be introduced in order to disclose whether a particular training record was used to train the model. 

In future studies, we will expand on this research and apply realistic attacks on AML detection algorithms in order to assess whether sensitive information can be disclosed or fool the model to hide illicit transactions. Based on that, we will also explore defence strategies to mitigate these attacks. One popular defence strategy against privacy attacks is differential privacy~\cite{abadi2016deep}. 
%\subsection{Future works: federated learning between banks}%draft section added by Paola 
Our main results learn from graphs which are totally available, such as graphs of banking transactions in blockchains where these are public. However, there are also many real-life scenario in classical banking where money-launderers take advantage of banks by not disclosing private transactions to split the laundered amounts into intermediates across different banks to go undetected (cf. the model of\cite{Soltani2016new}), and we would also like to provide tools to grant financial authorities the ability to detect such patterns with these privacy constraints, as a complementary approach to the more efficient learning methods on public graphs presented in previous sections. Work on this topic has already been done in literature~\cite{PP22}, to give a cryptographic functional encryption scheme to allow~\cite{Soltani2016new}'s similarity calculations to be performed by financial authorities on data from concurrent banks. However, size of embedding vectors on which the similarity scores are calculated are linear w.r.t number of existing bank accounts, which would be huge in a practical use-case. Using graph embedding techniques the way we have in previous sections could drastically improve these sizes.

One of the challenges in this context is for banks to provide embeddings of each of their nodes (the bank accounts), from the graph of transactions, knowing only the subgraph of the transactions made to and from the accounts they control. Because of this, we will perform an embedding depending only on nodes' first neighbors, those they are directly having transactions with. This is also relevant with respect to the corresponding first topological pattern in \cref{fig:ml_topologies}, as we are considering intermediate nodes split across different banks but having the same direct neighbors, and this is the sole criteria upon which we would want to have similar node embeddings. Hence, we use a variant of fast random projection embeddings that we describe with the following algorithm steps, to get the incoming (resp. outgoing) transaction embeddings: 

\begin{enumerate}%Paola: Maybe we could replace this with an algorithm-format figure
    \item Each node $n$ in the subgraph and its direct neighbors gets an initial random embedding vector value $e_{n,0}$ from a public hash function on the node, instantiated with a public seed (and with no null outputs). 
    \item For each node $n$ in the subgraph, denoting $N$ the set of its direct neighbors, and $w_{i\to j}$ the amounts sent from account $i$ to account $j$ (i.e. the weight on the oriented edge from $i$ to $j$, for any nodes $i$, $j$, and setting that weight to 0 when the nodes are not connected), $n$ get a new embedding vector: $e_{n,1} := \sum_{m\in N} e_{m,0} \cdot w_{m\to n}$ (and respectively, for the outgoing transaction embeddings: $e_{n,1} := \sum_{m\in N} e_{m,0} \cdot w_{n\to m}$). 
    \item The vectors for the previous step are then normalized; each node $n$ in the subgraph gets the embedding vector: $e_{n,3} := \frac{e_{n,2}}{||e_{n,2}||_2}$, where $||.||_2$ denotes the euclidian norm. This vector is then returned as the embedding vector $e_{n,\mathsf{in}}$ (resp. $e_{n,\mathsf{out}}$ for outgoing transactions).
\end{enumerate}

We then apply the inner-products and similarity calculations similar to literature~\cite{PP22,Soltani2016new} on these embedding vectors; to compare nodes $n$ from bank 1 and $m$ from bank 2, we compute: $\sigma(n,m) = \langle e_{n,\mathsf{in}};e_{m,\mathsf{in}} \rangle \times \langle e_{n,\mathsf{out}};e_{m,\mathsf{out}} \rangle$, and deem that nodes with high similarity values should be labeled as suspicious accounts which might be part of the same money-laundering network, having the bulk of their transactions from and to the same neighbors.

\section*{Acknowledgment} 
This work was supported by the BMBF-ANR funded project Crypto4Graph-AI. 

\bibliographystyle{IEEEtran}
\bibliography{references.bib}

\end{document}